# Comparison of computer systems and ranking criteria for automatic melanoma detection in dermoscopic images

Kajsa Møllersen[1]*, Maciel Zortea[2], Thomas R. Schopf[3], Herbert Kirchesch[4], Fred Godtliebsen[2]

**1** Department of Community Medicine, UiT The Arctic University of Norway, Tromsø, Norway, **2** Department of Mathematics and Statistics, UiT The Arctic University of Norway, Tromsø, Norway, **3** Norwegian Centre for E-health Research, University Hospital of North Norway, Tromsø, Norway, **4** Private office, Venloer Straße 107, 50259 Pulheim, Germany

* kajsa.mollersen@uit.no

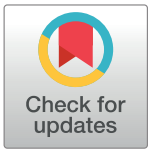

## Abstract

Melanoma is the deadliest form of skin cancer, and early detection is crucial for patient survival. Computer systems can assist in melanoma detection, but are not widespread in clinical practice. In 2016, an open challenge in classification of dermoscopic images of skin lesions was announced. A training set of 900 images with corresponding class labels and semi-automatic/manual segmentation masks was released for the challenge. An independent test set of 379 images, of which 75 were of melanomas, was used to rank the participants. This article demonstrates the impact of ranking criteria, segmentation method and classifier, and highlights the clinical perspective. We compare five different measures for diagnostic accuracy by analysing the resulting ranking of the computer systems in the challenge. Choice of performance measure had great impact on the ranking. Systems that were ranked among the top three for one measure, dropped to the bottom half when changing performance measure. Nevus Doctor, a computer system previously developed by the authors, was used to participate in the challenge, and investigate the impact of segmentation and classifier. The diagnostic accuracy when using an automatic versus the semi-automatic/manual segmentation is investigated. The unexpected small impact of segmentation method suggests that improvements of the automatic segmentation method w.r.t. resemblance to semi-automatic/manual segmentation will not improve diagnostic accuracy substantially. A small set of similar classification algorithms are used to investigate the impact of classifier on the diagnostic accuracy. The variability in diagnostic accuracy for different classifier algorithms was larger than the variability for segmentation methods, and suggests a focus for future investigations. From a clinical perspective, the misclassification of a melanoma as benign has far greater cost than the misclassification of a benign lesion. For computer systems to have clinical impact, their performance should be ranked by a high-sensitivity measure.

## Introduction

Melanoma is the deadliest form of skin cancer. The survival rate drops as the level of tumour invasion increases, and therefore early detection is crucial for the patient's survival [1, 2]. Melanoma detection relies mainly on visual inspection, which is challenging because melanomas often resemble benign skin lesions [3]. An aid in melanoma detection is the dermoscope, a simple device consisting of a magnifying lens and surrounding light, which is commonly used by dermatologists to examine skin conditions. The dermoscope reveals structures in lesions not visible to the naked eye, and has proven to increase diagnostic accuracy of skin lesions when used by trained personnel [4].

Computer systems for melanoma detection seek to correctly classify a lesion based on input and previous training. The ideal diagnostic test should have 100% sensitivity and 100% specificity. In medicine this is however almost never feasible. In the evaluation of clinical diagnostic aids for melanoma detection, the main priority is on increasing sensitivity as overlooking a melanoma may have a fatal outcome for the patient [5]. In contrast, classifying a benign lesion as melanoma usually implies an unnecessary excision of the skin lesion which some patients may find annoying but without any further consequences.

The input of the computer system is typically a dermoscopic image, but other technologies such as multispectral imaging and Rahman spectroscopy are also used. See e.g., [6] and [7] for an overview. Training relies on an available data set of labelled lesion images.

Research and development of computer systems for melanoma detection have increased in the past few decades, following the development of digital cameras and attachable dermoscopes. An overview can be found in [8]. Despite the effort made in developing new systems, their use in clinical practice is not widespread. An explanation may be the lack of convincing reports on the systems' diagnostic accuracy, owing to small data sets and non-comparable results.

Small data sets result in training and testing on the same data, often by incorrect use of cross-validation. This gives overly optimistic estimates of the test error [9, 10], and the classification accuracy of the system drops dramatically when tested on an independent test set. In the studies that report independent testing, the test set is often too small for the results to be generalised. In addition, consecutively collection of images is crucial for a clinical-like dataset. Only two studies that we know of have more than 50 melanomas in an independent test set of consecutively collected images [11, 12].

The diagnostic difficulty of the data set has significant impact on the reported diagnostic accuracy of the system, as discussed in [13], and illustrated in [12] and [14]. The diagnostic difficulty of a data set depends on factors such as proportion of melanomas, exclusion of non-melanocytic lesions, median Breslow depth, etc. Therefore, different systems cannot be compared unless they are tested on the same data set. There are few studies where several systems are tested on the same set of lesions, with the exception of [14] and [15].

A sufficiently large, publicly available data set makes independent testing and comparison of diagnostic accuracy between systems possible. In 2013, the PH$^2$ data set was made publicly available [16], and added important value to the research on computer systems for melanoma detection. Consisting of only 200 images, it does not overcome the aforementioned obstacles of small data sets. A larger data set, the ISIC Data Archive (see Section The challenge) is now publicly available. In 2016, an open challenge using images from the ISIC Data Archive was announced. Nevus Doctor, a computer system developed previously by the authors, was used to participate in this challenge. The design of the challenge allows for a direct comparison of diagnostic accuracy among the submitted melanoma detection systems, since they have all been tested on the same independent test set. However, there are unresolved obstacles of computer system performance ranking, of which this article shed light on two: the performance measure's impact on the ranking, and its clinical relevance.

Even when systems are tested on the same test set, the question on how to measure diagnostic accuracy arises. Summarising the performance of a system for the whole range of sensitivity and specificity values gives a different perspective than measuring the performance only for very high sensitivity values. Choice of measure is not trivial as different measures penalises misclassifications differently, and it can have huge impact on the ranking.

In addition, we investigate the potential impact of segmentation method and classifier through the variation of the diagnostic accuracy of Nevus Doctor. The rest of the article is organised as follows: Section *The challenge* describes the data, the challenge and the five performance measures. In addition, a brief description of Nevus Doctor is given. In Section *Results*, the variations in ranking of the top systems are presented. An investigation into segmentation method and classifier algorithm is also found here. A discussion of the results, both from a machine learning and clinical perspective, is provided in Section *Discussion*, together with some concluding remarks.

## The challenge

The International Skin Image Collaboration (ISIC) (http://isdis.net/isic-project/) has created a public repository for dermoscopic skin lesion images (https://isic-archive.com/). The current dataset (as of March 12, 2016) consists of 12,086 images and is by far the largest data set of dermoscopic images available. Each image is annotated as 'benign' or 'malignant' according to histopathological diagnosis or expert consensus based on revision of the image. The images come from different sources, and hence are taken with different cameras and dermoscopes.

As part of the International Symposium on Biomedical Imaging (ISBI) 2016, the open challenge *Skin Lesion Analysis towards Melanoma Detection* was hosted by ISIC. The challenge consisted of three sub challenges: Segmentation, Feature Extraction, and Classification. Details can be found in [17]. The Classification sub challenge had two parts: Automatic segmentation, where only the images were released, and Manual segmentation, where a semi-automatic or manually obtained mask was also released for each image. There were 25 participants in the Automatic segmentation part, and 18 participants in the Manual segmentation part. Many participated in both parts.

A set of 900 dermoscopic images of skin lesions with corresponding class labels and masks was available for training. The test set consisted of 379 images, of which 75 were of melanomas. All images are publicly available from the ISIC web site. To participate in the challenge, a posterior probability of malignancy for each image in the test set had to be submitted. To calculate the performance of the different systems, the following definitions were used: TP = true positive, FP = false positive, TN = true negative, and FN = false negative, where 'malignant' is positive and 'benign' is negative. Precision = $TP/(TP + FP)$, recall = sensitivity (SE) = $TP/(TP + FN)$, and specificity = $TN/(TN + FP)$. Two global measures of performance were calculated: area under the precision-recall curve (average precision), and area under the curve of the receiver-operating characteristic (AUC of ROC). In addition, specificity scores at sensitivities equal to 95%, 98% and 99% were calculated. We will refer to the latter ones as high-sensitivity measures.

## Nevus Doctor

Nevus Doctor is a computer system for detection of melanoma and non-melanocytic skin cancer in dermoscopic images. A detailed description can be found in [18] and [19]. The system consists of automatic segmentation, feature extraction, and classification.

For participation in the challenge, semi-automatic feature selection was conducted on the basis of the 900 training images, following the procedure described in [18]. Apart from the 59 features listed in [18], three new features, described in [20], were added to the feature pool. Two classifiers were considered: linear discriminant analysis (LDA) and quadratic discriminant analysis (QDA). The feature and classifier selection resulted in 16 features and the QDA classifier. Description of each feature can be found in [18] and in [19] (feature f1, f5, f6, f10, f11, f12, f16, f17, f18, f54, f55, f56, f57, and f58), and in [20] (feature f60 and f62).

The prior class probability in a discriminant analysis classifier can be estimated based on the class labels in the training set. The output of the classifier is the posterior probability of an observation belonging to a certain class. For calculation of sensitivity values from 0% to 100%, the threshold for the posterior probability is adjusted from 0 to 1. Another option is to adjust the prior probability for the class from 0 to 1. For this paper, we have recalculated the high-sensitivity measures by adjusting the prior probability to avoid the situation where several lesions have the saturated posterior probability of 1.00.

## Results

The participants were ranked according to diagnostic accuracy of the test set. Other aspects such as computation time were not considered. The challenge's ranking criterion was average precision, but scores for the other performance measures listed in Section *The challenge* were provided as well. This made it possible to rank the participants according to several criteria, not only the one chosen by the challenge. Average precision and AUC of ROC are global performance measures, from 0% sensitivity (no melanomas detected), to 100% sensitivity (all melanomas detected). The high-sensitivity measures provide performance measures given that a high proportion of the melanomas is detected.

For those participants that were ranked highest by one of the measures, Table 1 shows the rankings by the remaining measures. The score is given for the highest ranked participants. Descriptions of the algorithms used were not submitted for the challenge, and therefore the participants are only given a letter as identifier in the table. Participants B and D did not participate in the Manual segmentation part.

From Table 1, it is clear that choice of performance measure has strong impact on the resulting ranking. Participant A, which were ranked highest for both Automatic and Manual segmentation by the challenge, is ranked in the middle for the high-sensitivity measures. Participant C receives high rankings for high-sensitivity measures, but only medium rankings according to the challenge's measure.

### The impact of segmentation

From Table 1 we see that the ranking varies according to whether the participants use their own segmentation algorithm or the semi-automatic/manual segmentation provided by the

**Table 1. Rankings for those participants that were highest ranked by one measure.**

| Segmentation | Automatic (25 participants) | | | | | | | Manual (18 participants) | | | | |
|---|---|---|---|---|---|---|---|---|---|---|---|---|
| Participant | A | B | C | C | D | E | F | A | E | C | E | F |
| Av. precision | 0.64 | 3 | 11 | 11 | 2 | 6 | 8 | 0.62 | 3 | 7 | 3 | 2 |
| AUC of ROC | 3 | 0.83 | 6 | 6 | 4 | 8 | 5 | 4 | 0.81 | 5 | 1 | 2 |
| SE = 95% | 11 | 8 | 0.39 | 1 | 6 | 13 | 7 | 11 | 2 | 0.32 | 2 | 3 |
| SE = 98% | 15 | 14 | 1 | 0.33 | 4 | 13 | 5 | 11 | 1 | 5 | 0.29 | 2 |
| SE = 99% | 11 | 9 | 3 | 3 | 0.25 | 12 | 10 | 8 | 7 | 2 | 7 | 0.25 |

https://doi.org/10.1371/journal.pone.0190112.t001

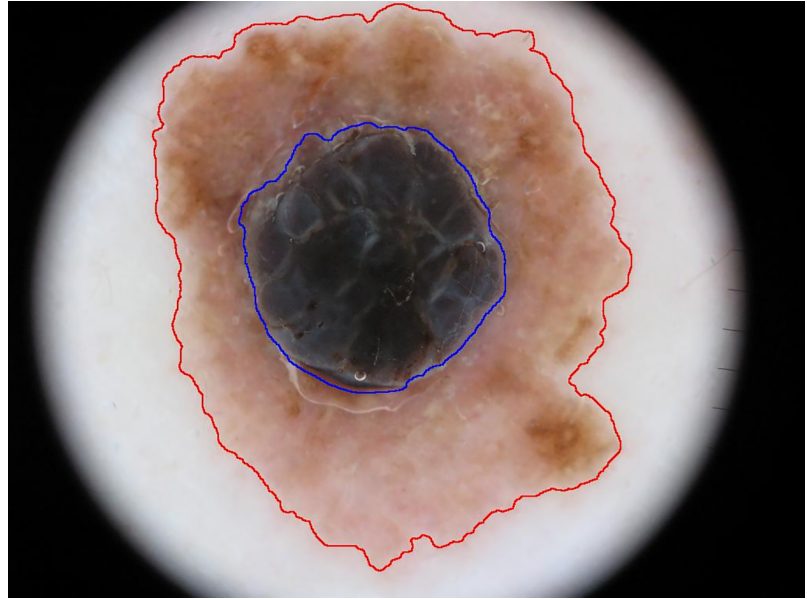

**Fig 1. Segmentation by different methods.** Blue line: automatic segmentation. Red line: manual segmentation. Image ISIC_000276 of the ISIC Archive, shared under the CC-0 license.

https://doi.org/10.1371/journal.pone.0190112.g001

challenge. Interestingly, the best score for each measure does not improve by semi-automatic/manual segmentation. This could mean that the automatic segmentation algorithms give approximately the same masks as the semi-automatic/manual segmentation, or it could mean that the segmentation mask does not influence the final classification significantly.

The algorithms for the challenge participants were not submitted, and therefore we investigated the impact of segmentation for Nevus Doctor. Fig 1 shows an example of the automatic segmentation used in Nevus Doctor and the manual segmentation provided by the challenge. In general, when the difference between the automatic and semi-automatic/manual segmentation was large, the automatic segmentation missed outer lighter areas of the lesion, like in Fig 1.

The occasionally failed segmentation does not seem to have an impact on the classification result. Fig 2 shows the ROC curves for automatic and semi-automatic/manual segmentation for Nevus Doctor. The two curves follow each other closely for the whole range of sensitivities/specificities. This is also confirmed by the results for the five performance measures, as seen in Table 2.

## The impact of classifier algorithm

Because segmentation method had lesser impact than expected, we investigated the impact of classifier algorithm. The performance of a classifier is tightly connected to the characteristics of the features, and it is therefore difficult to rank classifiers. The differences in performance can be the result of the features being more suited to a certain classifier, and the ranking might change if other features are made available.

We therefore investigate the performance of a few fairly similar classifiers. The aim is not to rank the classifiers for melanoma detection, but to investigate how the performance of a system changes by using a different classifier.

When training Nevus Doctor for the challenge, LDA and QDA classifiers were considered, and QDA was used on the test set. In addition, we have investigated the diagonal LDA and

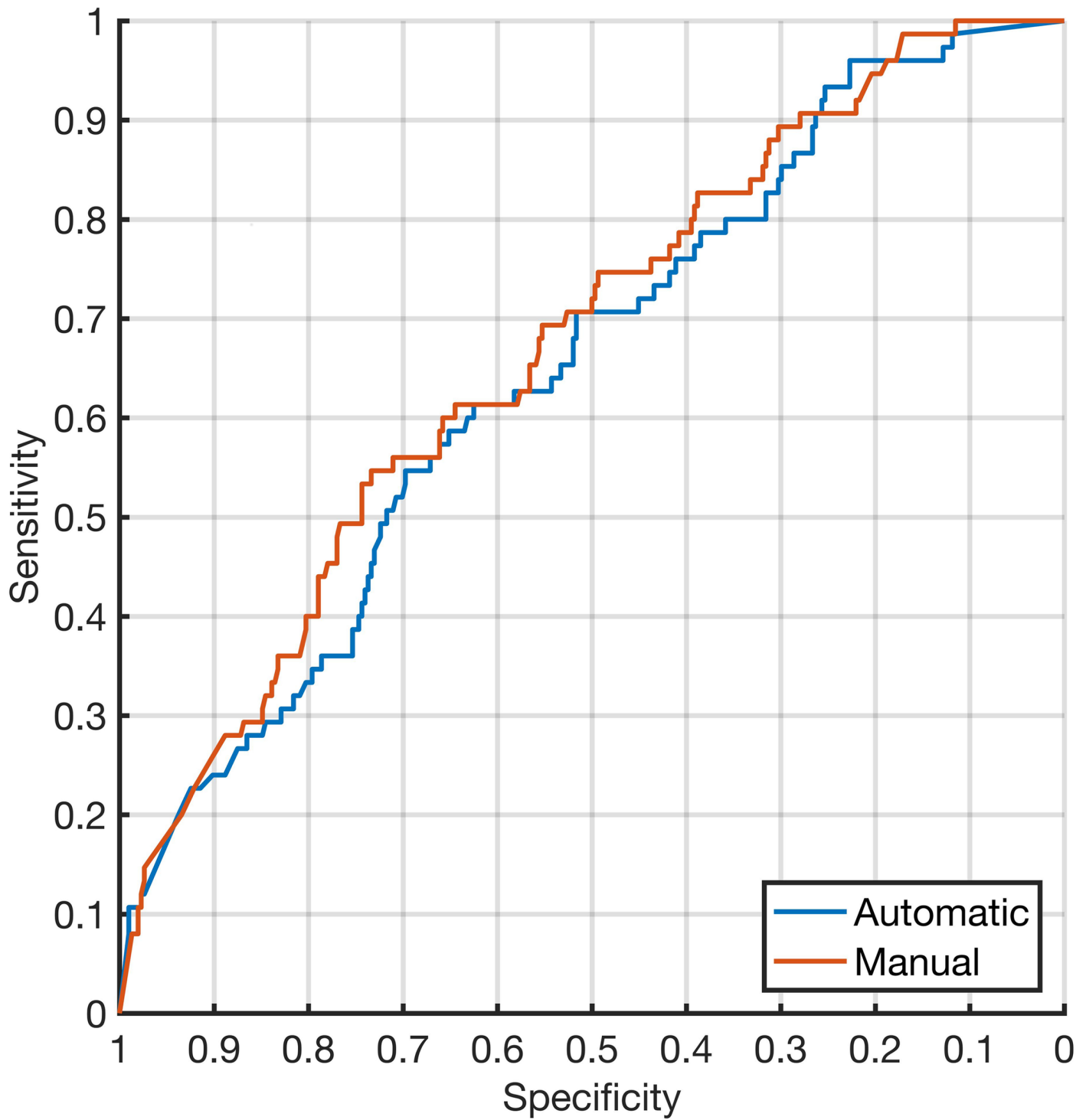

**Fig 2. ROC curves for different segmentation methods.** ROC curves for Nevus Doctor with automatic and semi-automatic/manual segmentation.

https://doi.org/10.1371/journal.pone.0190112.g002

QDA (dLDA and dQDA) classifiers, which correspond to naive Bayes classifiers. The features were not reselected, and therefore, QDA has a slight benefit. The results are shown in Table 3. We see that the performance varies considerably more between different classifiers than between the two segmentation methods.

Table 2. Scores for Nevus Doctor.

| Segmentation: | Automatic | Manual |
|---|---|---|
| Average precision | 0.35 | 0.37 |
| AUC of the ROC | 0.64 | 0.67 |
| SE = 95% | 0.23 | 0.19 |
| SE = 98% | 0.13 | 0.17 |
| SE = 99% | 0.12 | 0.17 |

https://doi.org/10.1371/journal.pone.0190112.t002

## Discussion

The ISIC data repository, consisting of more than 12,000 dermoscopic images, has answered to a need expressed for several years in the field of computer aided melanoma detection: a publicly available data set. Two important outcomes of collecting and sharing data are: (1) Large data sets allow for large training sets and independent test sets. (2) Public availability allows for comparison of algorithms on the same data. The ISBI challenge has taken advantage of this and set aside an independent test set, and ranked its participants according to diagnostic accuracy. However, we see from Table 1 that the choice of performance measure has great impact on the resulting ranking.

In this challenge, the systems were ranked according to average precision, which weights misclassifications equally whether it is a melanoma misclassified as benign, or a benign lesion misclassified as malignant. Specificity score at high sensitivity takes into account that a melanoma misclassified as benign can have fatal outcome for the patient, and has been used in clinical studies for melanoma detection [11, 12]. A problem with ranking by very high sensitivity is its dependence on the data set. In this challenge, 99% sensitivity corresponds to misclassifying only one melanoma, and under such circumstances, the ranking can be somewhat arbitrary. To avoid this, cross-validation, bootstrapping or similar methods can be used.

From Tables 1 and 2 and Fig 2 we see that whether automatic or semi-automatic/manual segmentation is used had little impact on the final classification accuracy. This implies that further improvement of automatic segmentation algorithms regarding resemblance to manual segmentation will not improve the classification accuracy. However, manual segmentation does not necessarily reflect the true borders of the lesion, which is a subjective opinion of experts, and hence, does not necessarily provide the best basis for correct classification. Further development of automatic segmentation algorithms should be aimed directly at classification performance, without the detour via manual segmentation, when the objective is increased diagnostic accuracy. But there is also a possibility that further development will have little impact and that resources should be allocated elsewhere.

In Table 3, fairly similar classifiers, using the same set of features, give more variation in the classification accuracy than different segmentation methods. Although this does not imply that other systems have the same variation in performance, it is an indication that choice of classifier might have bigger impact than suggested earlier [21].

Table 3. High-sensitivity measures for Nevus Doctor using different classifiers.

| | Automatic segmentation | | | |
|---|---|---|---|---|
| | LDA | QDA | dLDA | dQDA |
| SE = 95% | 0.14 | 0.23 | 0.28 | 0.32 |
| SE = 98% | 0.07 | 0.13 | 0.22 | 0.27 |
| SE = 99% | 0.05 | 0.12 | 0.21 | 0.24 |

https://doi.org/10.1371/journal.pone.0190112.t003

An important question when analysing the diagnostic accuracy, whatever performance measure, is its generalisability to other data sets. Ideally, the results should reflect a system's performance under actual clinical conditions. The test set in this challenge was not consecutively collected, and inclusion and exclusion criteria are not stated (e.g. if non-melanoma skin cancers are excluded). Even though the performance in absolute numbers, e.g. specificity at 95% sensitivity, are better for many of the systems in this challenge compared to other studies, one cannot draw any conclusions.

Demanding more information can result in fewer deposits to the repository, but to have clinical impact, a minimum of clinical information is essential. See e.g., [22] for recommendations.

The ISIC Data Archive consists of images taken with different cameras and dermoscopes, and from different populations, unlike many commercial systems that require the use of a certain camera and dermoscope. Robustness to variability in equipment is beneficial, due to the fast development of camera and dermoscope technologies, giving the system the ability to easily adapt to new equipment, without an extensive training set. However, it introduces some obstacles because the size and colour of a lesion have diagnostic importance, but are not consistent between different cameras and dermoscopes. Different image resolutions, magnifications and distances to the skin result in the true size of the lesion differing from image to image, though this can be solved easily by providing information regarding pixels per mm. Additional aspects that pose a challenge are the different colour calibrations of the cameras and different light sources, as illustrated in Fig 3. Colour is one of the main clinical features for melanoma detection [5], and computer systems normally have one or more colour features [23]. There have been attempts at automatic colour calibration based on image content [24], but the impact on classification is not fully investigated. A system's robustness can be of great importance in clinical practice, since a camera's colour calibration drifts over time, and the temperature of the diodes and the remaining voltage of the batteries influences the emitted light. On the other hand, a computer system that is robust to variability in equipment cannot be fine tuned and loses some of its classification accuracy [25]. The variability in the ISIC Data Archive is greater than within-equipment variability, but adds the possibility for robustness measures.

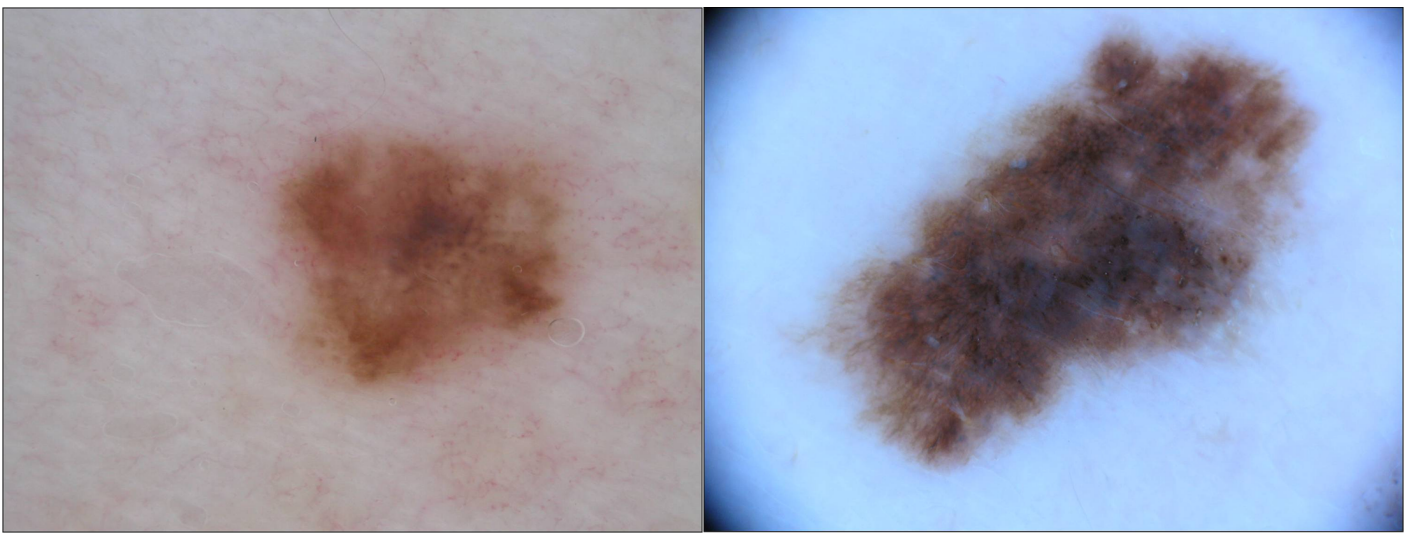

**Fig 3. Different colour.** Example of different colour calibration and/or different light source from the ISIC Data Archive. Image ISIC_0000023 and ISIC_0000037 of the ISIC Archive, shared under the CC-0 license.

https://doi.org/10.1371/journal.pone.0190112.g003

For the recent years, deep convolutional neural networks (CNN) have entered the field of image classification with great success, and has also been applied to melanoma detection [26]. In this study, CNN's performance was comparable to that of dermatologists, which is the same as reported for classical melanoma detection systems. A direct comparison between methods can give an answer to whether one set of methods outperforms the other. The ISBI 2017 melanoma detection challenge has added an additional requirement of submitting a description of the algorithms used, and a comparison between methods can be done once the results are published.

Computer systems for melanoma detection have considerably lower diagnostic accuracy than the current gold standard (histopathology), and can only provide decision support, not actual diagnosis. The usefulness of a system relies not only on its ability for correct classification but also on the possibility for the user to take advantage of the output. To truly know whether a system can have clinical impact, it must be tested in a clinical setting, which is not feasible for an open challenge.

## Conclusion

The *Skin Lesion Analysis towards Melanoma Detection* challenge has resulted in the largest comparison between computer systems for dermoscopic images, both in number of participants and in test set size. Whereas the obstacle of small datasets and non-comparable results are solved with a public repository and an open challenge, the question of performance measure still remains. This article has investigated accuracy measures' impact on classification performance ranking, and has thus highlighted a choice that is seldom under careful consideration. From a clinical perspective, the misclassification of a melanoma as benign has far greater cost than the misclassification of a benign lesion as malignant. A system with 95% sensitivity misclassifies one in 20 melanomas, and it is difficult to imagine that lower sensitivity is acceptable, and global measures of accuracy are not clinically relevant. We suggest a specificity score for a fixed sensitivity of at least 95%, calculated by the use of cross-validation, alternatively the AUC of ROC calculated from 95% to 100% sensitivity, also using cross-validation. A high-sensitivity measure can improve the clinical relevance of the challenge, and help ISIC to achieve its goal "to help reduce melanoma mortality" (http://isdis.net/isic-project/).

## Acknowledgments

We wish to thank the reviewers for valuable feedback which has improved the manuscript.

## Author Contributions

**Conceptualization:** Kajsa Møllersen, Maciel Zortea, Fred Godtliebsen.

**Data curation:** Kajsa Møllersen, Maciel Zortea.

**Formal analysis:** Kajsa Møllersen, Maciel Zortea.

**Investigation:** Maciel Zortea.

**Software:** Kajsa Møllersen, Maciel Zortea.

**Validation:** Kajsa Møllersen.

**Visualization:** Kajsa Møllersen, Maciel Zortea.

**Writing – original draft:** Kajsa Møllersen.

**Writing – review & editing:** Kajsa Møllersen, Maciel Zortea, Thomas R. Schopf, Herbert Kirchesch, Fred Godtliebsen.

## References

1. American Cancer Society. Cancer Facts & Figures 2016. American Cancer Society; 2016.
2. Cancer Registry of Norway. Cancer in Norway 2015—Cancer incidence, mortality, survival and prevalence in Norway. Cancer Registry of Norway; 2016.
3. Strayer SM, Reynolds PL. Diagnosing skin malignancy: assessment of predictive clinical criteria and risk factors. The Journal of family practice. 2003; 52(3):210–218. PMID: 12620175
4. Kittler H, Pehamberger H, Wolff K, Binder M. Diagnostic accuracy of dermoscopy. The Lancet Oncology. 2002; 3(3):159–165. https://doi.org/10.1016/S1470-2045(02)00679-4 PMID: 11902502
5. Argenziano G, Soyer HP, Chimenti S, Talamini R, Corona R, Sera F, et al. Dermoscopy of pigmented skin lesions: Results of a consensus meeting via the Internet. Journal of the American Academy of Dermatology. 2003; 48(5):679–693. https://doi.org/10.1067/mjd.2003.281 PMID: 12734496
6. Fink C, Haenssle HA. Non-invasive tools for the diagnosis of cutaneous melanoma. Skin Res Technol. 2016; 261–271. PMID: 27878858
7. Smith L, MacNeil S. State of the art in non-invasive imaging of cutaneous melanoma. Skin Research and Technology. 2011; 17(3):257–269. https://doi.org/10.1111/j.1600-0846.2011.00503.x PMID: 21342292
8. Korotkov K, Garcia R. Computerized analysis of pigmented skin lesions: A review. Artificial Intelligence in Medicine. 2012; 56(2):69–90. https://doi.org/10.1016/j.artmed.2012.08.002 PMID: 23063256
9. Smialowski P, Frishman D, Kramer S. Pitfalls of supervised feature selection. Bioinformatics. 2010; 26(3):440–443. https://doi.org/10.1093/bioinformatics/btp621 PMID: 19880370
10. Hastie T, Tibshirani R, Friedman J. The Elements of Statistical Learning: Data mining, Inference, and Prediction. 2nd ed. Springer Series in Statistics. New York: Springer; 2009. Available from: http://statweb.stanford.edu/~tibs/ElemStatLearn/.
11. Monheit G, Cognetta AB, Ferris L, Rabinovitz H, Gross K, Martini M, et al. The Performance of MelaFind: A Prospective Multicenter Study. Archives of Dermatology. 2011; 147(2):188–194. https://doi.org/10.1001/archdermatol.2010.302 PMID: 20956633
12. Malvehy J, Hauschild A, Curiel-Lewandrowski C, Mohr P, Hofmann-Wellenhof R, Motley R, et al. Clinical performance of the Nevisense system in cutaneous melanoma detection: An international, multicentre, prospective and blinded clinical trial on efficacy and safety. The British journal of dermatology. 2014; 171(5):1099–1107. https://doi.org/10.1111/bjd.13121 PMID: 24841846
13. Rosado B, Menzies S, Harbauer A, Pehamberger H, Wolff K, Binder M, et al. Accuracy of Computer Diagnosis of Melanoma: A Quantitative Meta-analysis. Archives of Dermatology. 2003; 139(3):361–367. https://doi.org/10.1001/archderm.139.3.361 PMID: 12622631
14. Møllersen K, Kirchesch H, Zortea M, Schopf TR, Hindberg K, Godtliebsen F. Computer-Aided Decision Support for Melanoma Detection Applied on Melanocytic and Nonmelanocytic Skin Lesions: A Comparison of Two Systems Based on Automatic Analysis of Dermoscopic Images. BioMed Research International. 2015; 2015:1–8.
15. Perrinaud A, Gaide O, French LE, Saurat JHH, Marghoob AA, Braun RP. Can automated dermoscopy image analysis instruments provide added benefit for the dermatologist? A study comparing the results of three systems. The British journal of dermatology. 2007; 157(5):926–933. https://doi.org/10.1111/j.1365-2133.2007.08168.x PMID: 17854361
16. Mendonca T, Ferreira PM, Marques JS, Marcal ARS, Rozeira J. PH[2]—A dermoscopic image database for research and benchmarking. In: Engineering in Medicine and Biology Society (EMBC), 2013 35th Annual International Conference of the IEEE. IEEE; 2013. p. 5437–5440.
17. Gutman D, Codella NCF, Celebi E, Helba B, Marchetti M, Mishra N, et al. Skin Lesion Analysis toward Melanoma Detection: A Challenge at the International Symposium on Biomedical Imaging (ISBI) 2016, hosted by the International Skin Imaging Collaboration (ISIC); 2016. Available from: http://arxiv.org/abs/1605.01397.
18. Møllersen K, Zortea M, Hindberg K, Schopf TR, Skrøvseth SO, Godtliebsen F. Improved skin lesion diagnostics for general practice by computer aided diagnostics. In: Celebi ME, Mendonca T, Marques JS, editors. Dermoscopy Image Analysis. CRC Press; 2015. p. 247–292.
19. Zortea M, Schopf TR, Thon K, Geilhufe M, Hindberg K, Kirchesch H, et al. Performance of a Dermoscopy-based Computer Vision System for the Diagnosis of Pigmented Skin Lesions Compared with Visual Evaluation by Experienced Dermatologists. Artificial Intelligence in Medicine. 2014; 60(1):13–26. https://doi.org/10.1016/j.artmed.2013.11.006 PMID: 24382424
20. Møllersen K, Hardeberg JY, Godtliebsen F. Divergence-based colour features for melanoma detection. In: Colour and Visual Computing Symposium (CVCS), 2015. IEEE; 2015. p. 1–6.

21. Maglogiannis I, Doukas CN. Overview of Advanced Computer Vision Systems for Skin Lesions Characterization. IEEE Transactions on Information Technology in Biomedicine. 2009; 13(5):721–733. https://doi.org/10.1109/TITB.2009.2017529 PMID: 19304487

22. Malvehy J, Puig S, Argenziano G, Marghoob AA, Soyer PP, International Dermoscopy Society Board members. Dermoscopy report: Proposal for standardization. Results of a consensus meeting of the International Dermoscopy Society. Journal of the American Academy of Dermatology. 2007; 57(1):84–95. https://doi.org/10.1016/j.jaad.2006.02.051 PMID: 17482314

23. Madooei A, Drew MS. Incorporating Colour Information for Computer-Aided Diagnosis of Melanoma from Dermoscopy Images: A Retrospective Survey and Critical Analysis. International Journal of Biomedical Imaging. 2016; 2016. https://doi.org/10.1155/2016/4868305 PMID: 28096807

24. Barata C, Celebi ME, Marques JS. Improving Dermoscopy Image Classification Using Color Constancy. IEEE Journal of Biomedical and Health Informatics. 2015; 19(3):1146–1152. PMID: 25073179

25. Wolpert DH, Macready WG. No free lunch theorems for optimization. Evolutionary Computation, IEEE Transactions on. 1997; 1(1):67–82. https://doi.org/10.1109/4235.585893

26. Esteva A, Kuprel B, Novoa RA, Ko J, Swetter SM, Blau HM, et al. Dermatologist-level classification of skin cancer with deep neural networks. Nature. 2017; 542(7639):115–118. https://doi.org/10.1038/nature21056 PMID: 28117445